\documentclass{article}
\usepackage{ijcai21}

\usepackage{times}
\usepackage{soul}
\usepackage{url}
\usepackage[hidelinks]{hyperref}
\usepackage[utf8]{inputenc}
\usepackage[small]{caption}
\usepackage{graphicx}
\usepackage{amsmath}
\usepackage{amsthm}
\usepackage{booktabs}
\usepackage{algorithmic}
\usepackage{epsfig}
\usepackage{amssymb}
\usepackage{dsfont}
\usepackage{dsfont}
\usepackage{booktabs}
\usepackage{pifont}
\usepackage{arydshln}
\usepackage{multirow}
\usepackage{mathrsfs}
\usepackage{mathtools}
\usepackage[ruled]{algorithm2e}

\SetKwComment{Comment}{$\triangleright$\ }{}

\usepackage{xcolor}

\usepackage{soul}

\newcommand{\ie}{\textit{i}.\textit{e}.\ }
\newcommand{\eg}{\textit{e}.\textit{g}.\ }

\DeclareMathOperator*{\argmin}{arg\,min}

\title{A Survey on Universal Adversarial Attack}

\author{
    Chaoning Zhang$^{1*}$\and
    Philipp Benz$^{1}$\footnote{Equal Contribution}\and
    Chenguo Lin$^{2*}$\and
    Adil Karjauv$^1$\and 
    Jing Wu$^3$\and
    In So Kweon$^1$\
    \affiliations
    $^{1}$Korea Advanced Institute of Science and Technology\\
    $^{2}$Sichuan University\\
    $^3$University of Electronic Science and Technology of China\\
    \emails
    chaoningzhang1990@gmail.com, 
    pbenz@kaist.ac.kr, 
    linchenguo@stu.scu.edu.cn\\
    mikolez@gmail.com,
    wujing@std.uestc.edu.cn,
    iskweon77@kaist.ac.kr
}

\begin{document}

\maketitle

\begin{abstract}
The intriguing phenomenon of adversarial examples has attracted significant attention in machine learning and what might be more surprising to the community is the existence of universal adversarial perturbations (UAPs), \ie a single perturbation to fool the target DNN for most images. With the focus on UAP against deep classifiers, this survey summarizes the recent progress on universal adversarial attacks, discussing the challenges from both the attack and defense sides, as well as the reason for the existence of UAP. We aim to extend this work as a \textbf{dynamic} survey that will \textbf{regularly update its content} to follow new works regarding UAP or universal attack in a wide range of domains, such as image, audio, video, text, etc. Relevant updates will be discussed at: \url{https://bit.ly/2SbQlLG}. We welcome authors of future works in this field to contact us for including your new findings. 
\end{abstract}










\section{Introduction}

Deep neural networks (DNNs) have achieved milestone performances in numerous computer vision tasks. However, despite their success, DNNs have been discovered to be vulnerable to adversarial examples~\cite{szegedy2013intriguing}, carefully crafted, quasi-imperceptible perturbations, which fool a DNN when added to an image. More interestingly, the existence of image-agnostic (universal) adversarial perturbations has been shown in recent works. A universal adversarial perturbation (UAP) is a single perturbation that is capable to fool a DNN when added to most natural images~\cite{moosavi2017universal}. The discovery of UAPs led to various explorations of this phenomenon, \eg universal adversarial attack, the defense against UAPs as well as attempts to understand the phenomenon of UAPs. Even though UAPs have initially been studied in the domain of image classification, their exploration has expanded into other domains as well. 

\paragraph{Scope of the survey.}
To this date, the amount of works on adversarial robustness is so large that it is impossible to cover them in a single survey. We refer the readers to~\cite{akhtar2018threat} for an introduction to general adversarial attack and defense. 
The focus of this survey is mainly on the advancements on a special type of adversarial attack, \ie universal adversarial attack, in the last few years. 
It is worth mentioning that image classification is the main application field where researchers design new attack and defense techniques and analyze adversarial perturbations. The core element of universal attack lies in the UAP, which can be generated beforehand and then directly applied with a simple summation operation during the attack stage. In this work, unless specified, we discuss the UAP in the context of image classification. \textit{We highlight that this work will be extended as a dynamic survey that will update its content for including new works in this field and any feedback is welcome.} 

\paragraph{Structure.} The survey is structured as follows: First, the basic notion and notation of UAPs in the context of image-classification will be introduced. Then universal adversarial attack methods will be covered, followed by defense methods against UAPs. Afterward, an overview will be given about the different perspectives on the understanding of the UAP phenomenon. We further identify data-dependency, black-box attack capabilities, and class-discrimination as three challenges of UAPs and discuss them. Finally, works covering UAPs going beyond image-classification will be discussed.

\section{A Short Primer on Image-Dependent Attack Methods} 
Before we get into the topic of UAP, it is relevant to discuss general image-dependent adversarial attacks since most UAP algorithms are developed based on image-dependent attacks. We categorize the adversarial attack into two groups: (a) minimizing perturbation magnitude given that the image is misclassified; (b) maximizing the attack success rate given a limited perturbation budget. \citeauthor{szegedy2013intriguing} proposed the first adversarial attack algorithm, box-constrained L-BFGS, to generate perturbations that can fool a network~\cite{szegedy2013intriguing}. This algorithm falls into the group (a). Another popular attack method is the Carlini and Wagner (C\&W) attack~\cite{carlini2017towards}. In essence, the C\&W attack is the same as the L-BFGS attack, but with a different loss function applied. \citeauthor{carlini2017towards} investigate multiple loss functions and find that the loss that maximizes the gap between the target class logit and highest logit (excluding the target class logit) results in superior performance. Yet another popular attack falling into this group is DeepFool that crafts perturbations iteratively by updating the gradient with respect to the model’s decision boundaries. In every iteration, DeepFool chooses the perturbation direction of the minimum magnitude that is orthogonal to the decision hyperplane. With the goal of finding pseudo-minimal perturbation, group (a) has the disadvantage of being cumbersome and slow. Relatively, Group (b) that maximizes the attack success rate given a limited budget is more straightforward. The first algorithm that falls into this group is the Fast Gradient Sign Method (FGSM)~\cite{goodfellow2014explaining}. FGSM is simple and fast, which comes at the cost of its effectiveness. Iterative FGSM (I-FGSM)~\cite{kurakin2016adversarial}, iteratively performs the FGSM attack. In each iteration, only a fraction of the allowed noise limit is added, which contributes to its higher attack effect compared to FGSM. Another widely used white-box attack method is termed PGD introduced in~\cite{madry2017towards}. In essence, PGD is the same as I-FGSM and the only difference lies in that the PGD attack initializes the perturbation with random noise while I-FGSM just initializes the perturbation with zero values. This random initialization can help improve the attack success rate, especially when the number of iterations is limited to a relatively small value. Another advantage of the initialization is that it can further help improve the attack success rate with multiple trials. 



\section{Image-Agnostic Adversarial Attacks}

\subsection{Definition of UAPs in Deep Image Classifiers}
\label{sec:uap_def}
The existence of UAPs to fool the deep classifier for most images has first been demonstrated in~\cite{moosavi2017universal}, and we will mainly follow the notation introduced in their work. Given a distribution of images $\mu$ in $\mathds{R}^d$ and a classifier function $\hat{k}$, we denote the output of the classifier given an image $x \in \mathds{R}^d$ as $y=\hat{k}(x)$. The overall objective is to find a single perturbation vector $\nu \in \mathds{R}^d$, such that the $\hat{k}$ is fooled for most encountered images. Additionally, $\nu$ should be sufficiently small, which is commonly modeled through an upper-bound $\epsilon$ on the $l_p$-norm, commonly denoted as $||\cdot||_p$, of the perturbation, \ie $||\nu||_p \leq \epsilon$. More formally, we seek a UAP, \ie $\nu$ such that:
\begin{equation}
    \hat{k}(x + v) \neq \hat{k}(x) \text{ for most $x\sim \mu$} \quad \text{s.t.} \quad ||\nu||_p \leq \epsilon.
    \label{eq:uap_objective}
\end{equation}
A popular choice is to set $p=\infty$, and to set the value of $\epsilon$ to $10/255$, assuming images to be in the range $[0,1]$~\cite{moosavi2017universal,poursaeed2018generative,zhang2020understanding}. 
 
\subsection{Metrics to Evaluate the UAP Effectiveness} Given the above definition of UAPs, the \textit{fooling ratio} is the most widely adopted metric for evaluating the efficacy of the generated UAP. Specifically, the \textit{fooling ratio} is defined as the percentage of samples whose prediction changes after the UAP is applied, \ie $\underset{x\sim X}{\rm I\!P}(\hat{k}(x + \nu) \neq \hat{k}(x))$. Some works~\cite{zhang2020understanding,benz2020double} have investigated targeted UAPs whose goal is to flip the prediction of most samples to a pre-defined target class. The \textit{targeted fooling ratio} is defined as $\underset{x\sim X}{\rm I\!P}(f(x + \nu) = t)$, where $t$ is the target label.  

\subsection{Universal Attack Methods}
\label{sec:attack_methods}

\paragraph{The vanilla universal attack.} 
UAPs were first introduced in~\cite{moosavi2017universal}. 
The proposed algorithm accumulates the UAP by iteratively crafting image-dependant perturbations for the data points. Specifically, if the already accumulated perturbation does not send the current data point across the decision boundary, the minimal perturbation $\Delta \nu$ is computed to send the sample over the decision boundary. After every iteration update, the perturbation is projected on the $l_p$ ball of radius $\epsilon$. In the vanilla UAP algorithm the projection operator $\mathcal{P}_{p, \epsilon}$ is defined as:
$\mathcal{P}_{p, \epsilon} = \argmin_{\nu'}||\nu -\nu'|| \text{ subject to } ||\nu'||_p \leq \epsilon$.
The accumulation of minimal perturbations is repeated until the fooling rate of $\nu$ exceeds a certain threshold. The authors note that the number of encountered data points can be smaller than the number of total training points.


\paragraph{Generating UAPs with singular vectors (SV-UAP).} 
A different algorithm to craft UAPs has been proposed in~\cite{khrulkov2018art}. Their method is based on the calculation of the singular vectors of the Jacobian matrices of the feature maps to obtain UAPs. The proposed approach shows a good data-efficiency, which can generate UAPs with a fooling rate of more than $60\%$ on the ImageNet validation set by using only 64 images.

\paragraph{Generating UAPs with generative networks.} 
A Network for adversary generation \textbf{(NAG)} was first introduced by~\cite{mopuri2018nag}. Inspired by Generative Adversarial Networks (GAN)~\cite{goodfellow2014generative}, NAG aims to model the distribution of UAPs. Therefore the authors modify the GAN framework by replacing the discriminator with the (frozen) target model and introduce a novel loss to train the generator. The novel loss function is composed of a fooling objective and a diversity objective. As the name suggests, the fooling objective is designed such that the generated perturbation fools the target classifier. Specifically, the loss is formulated to encourage the generator to generate perturbations that decrease the confidence of the original (benign) predictions. The diversity objective encourages the diversity of perturbations by increasing the distance of their feature embeddings predicted by the target classifier.
Another variant of generative adversarial perturbations \textbf{(GAP)} using a generator to craft UAPs was also explored in~\cite{poursaeed2018generative}. 
The objective is to train a generative network that transforms a random pattern to an image-dependant perturbation or UAP. The scale operation is introduced to guarantee the perturbation lies in a certain range. Concurrent to this, the authors of ~\cite{hayes2018learning} also explored the idea of generating adversarial perturbations with a generator network. 

\paragraph{Dominant Feature-UAP (DF-UAP).} \cite{zhang2020understanding} treats the UAP as network weights and apply the DNN training techniques, such as Adam optimizer and batch training, to maximize feature content of a target class. In both non-targeted and targeted setting, the resultant UAP has dominant features (DF). \citeauthor{zhang2020understanding} investigate various loss functions in the context of targeted UAP generation. For the non-targeted setting, \citeauthor{zhang2020understanding} further propose a cosine similarity based loss for alleviating the need of ground-truth labels. 

\paragraph{A comparison between the different algorithms.}
The vanilla algorithm~\cite{moosavi2017universal} attacks a single image at once, scaling the number of iterations linearly with the number of processed images, leading to slow convergence. Moreover, their algorithm is based on the image-dependant DeepFool attack~\cite{moosavi2016deepfool}, which is overall found to be one of the slower attack techniques. \citeauthor{dai2019fast} identify that minimum perturbation resulted from the DeepFool is not optimal for the efficient UAP generation. At each iteration, instead of choosing the minimal perturbation vector, they proposed to choose the perturbation that has a similar orientation to the former one. Their empirical results demonstrate that this technique can help boost both the convergence and performance, leading to an increase of $9\%$ in fooling rate over the vanilla UAP attack. 
The generative networks-based approaches somewhat alleviate the rather cumbersome and slow procedure of the vanilla UAP algorithm. Adopting generative networks has the benefit that conventional training techniques can be applied to obtain a powerful UAP generation network, which overall showed superior performance over the early UAP generation methods. However, the requirement of a generative model itself is a drawback of these UAP generation approaches. The simple methods which directly update the perturbation with the calculated gradient proposed in~\cite{zhang2020understanding,shafahi2020universal,zhang2021jigsaw} demonstrate that a direct optimization of the UAP does not only remove the requirement to train a separate generative network but can also achieve superior performance. We provide an overview of different UAP generation methods in the white-box attack scenario in Table~\ref{tab:white_box_performance_comparison} supporting the here presented discussion with quantitative results.

\begin{table}[t]
\centering
    \small
    \setlength\tabcolsep{1.2pt}
    \scalebox{0.73}{
    \begin{tabular}{cccccc}
        \toprule
        Method  & AlexNet & GoogleNet  & VGG16  & VGG19  & ResNet152 \\
        \midrule
        UAP~\cite{moosavi2017universal}       & $93.3$           & $78.9$           & $78.3$           & $77.8$           & $84.0$           \\
        SV-UAP~\cite{khrulkov2018art} & $-$ & $-$ & $52.0$ & $60.0$ & $-$ \\
        GAP~\cite{poursaeed2018generative}    &  -               &  $82.7$          & $83.7$           & $80.1$           &     -       \\
        NAG~\cite{mopuri2018nag} & $96.44$ & $90.37$ & $77.57$ & $83.78$ & $87.24$ \\
        DF-UAP~\cite{zhang2020understanding} & $96.17$ & $88.94$ & $94.30$ & $94.98$ & $90.08$ \\
        Cos-UAP~\cite{zhang2021jigsaw} & $96.5$ & $90.5$ & $97.4$ & $96.4$ & $90.2$\\
        \midrule
        FFF~\cite{Mopuri2017datafree}         & $80.92$          & $56.44$          & $47.10$          & $43.62$          & -                \\
        AAA~\cite{mopuri2018ask}               & $89.04$          & $75.28$          & $71.59$          & $72.84$          & $60.72$          \\
        GD-UAP~\cite{mopuri2018generalizable} & $87.02$          & $71.44$          & $63.08$          & $64.67$          & $37.3$           \\
        PD-UA~\cite{liu2019universal} & $-$ & $67.12$ & $53.09$ & $48.95$ & $53.51$ \\
        DF-UAP (COCO)~\cite{zhang2020understanding} & $89.9$ & $76.8$ & $92.2$ & $91.6$ & $79.9$ \\
        Cos-UAP (Jigsaw)~\cite{zhang2021jigsaw} & $91.07$ & $87.57$ & $89.48$ & $86.81$ & $65.35$ \\
        \bottomrule
    \end{tabular}
    }
    \caption{Fooling ratio ($\%$) of different UAP generation methods in the white-box attack scenario. The results are divided into universal attacks with access to the original ImageNet training data (upper) and data-free methods (lower).} 
    \label{tab:white_box_performance_comparison}
\end{table}

\subsection{Defending Against UAP}
To mitigate the effect of adversarial perturbations, numerous works have attempted to either detect or defend through various techniques. To our knowledge, adversarial learning is the only defense method that has not been broken by strong white-box attacks~\cite{madry2017towards}, thus it has become the de-facto most widely used defense technique. A wide range of works~\cite{goodfellow2014explaining,madry2017towards,shafahi2019adversarial,zhang2019you,wong2020fast} have investigated adversarial training, but the scope of these techniques is often limited to image-dependent attacks. Here, we summarize relevant advancements on defending against UAPs. One straightforward approach to extend adversarial training to the field of universal attack is to replace the image-dependent adversarial examples with the samples perturbed by the UAP during network training. The main challenge lies in the fact that an effective UAP often takes many iterations to converge, thus adversarial training against universal attacks is challenging in practice due to constraints in computation resources. Note that it can be (N+1) time slower than normal training, where N is the required number of attack iterations. To address this concern,~\cite{moosavi2017universal} proposes to fine-tune the model parameters with the images perturbed by pre-computed UAPs. Unfortunately, this only leads to marginal robustness enhancements against UAPs, which is somewhat reasonable because the pre-computed fixed UAP is unlike the dynamically generated perturbation for normal (image-dependent) adversarial training. Thus, the model would be expected to be only robust to the fixed perturbations. To alleviate such concern, \citeauthor{mummadi2019defending} have proposed to generate UAPs on-the-fly through shared adversarial training~\cite{mummadi2019defending}. However, it still takes 20 times more computation resources than the normal training because the UAP generation process resembles the multi-step PGD adversarial training~\cite{madry2017towards}. Universal adversarial training (UAT)~\cite{shafahi2020universal} elegantly handles this issue by concurrently updating the networks and perturbation, resulting in fast adversarial training~\cite{wong2020fast}. Identifying that the UAP does not attack all classes equally, a recent work~\cite{benz2021universal} extends the UAT with class-wise perturbations, enhancing the robustness against the attack of UAP by a large margin. Moreover, it also leads to a more balanced class-wise robustness against UAP. 
The adversarial training on UAP has been perceived as a two-player zero-sum game~\cite{perolat2018playing}. 
Beyond adversarial training, a defense against UAPs has also been applied on the feature-level, through selective feature generation in~\cite{borkar2020defending}. Another framework for defending against UAP is proposed in~\cite{akhtar2018defense} which has two components: (a) Perturbation Rectifying Network (PRN) used as a rectifier to de-noise the UAP in the adversarial examples; (b) a binary classifier that detects adversarial examples perturbed through UAPs. 

\section{On the Existence of UAP}

The fundamental reason that adversarial examples are intriguing to the community is that a well-trained deep classifier can be fooled by a small imperceptible perturbation. It is counter-intuitive that human invisible adversarial perturbation can fool the target model, which motivates numerous works attempting to explain its existence from a wide range of perspectives, such as the local linearity of DNNs  ~\cite{goodfellow2014explaining}, input high-dimensionality~\cite{shafahi2018adversarial} and over-fitting~\cite{schmidt2018adversarially}, and noise disturbance~\cite{fawzi2016robustness}. 
Those explanations are limited to explain only image-dependent perturbations, in other words, they can not be easily extended to explain the image-agnostic properties of UAPs. The investigation on the existence of UAP is still in its infancy and in the following, we summarize the works in the literature on the existence of UAP. Specifically, we find that those explanations can be divided into two categories: (a) geometric perspective; (b) feature perspective.

\paragraph{Understanding UAPs from a geometric perspective.} 
\citeauthor{moosavi2017universal} have attributed the existence of UAPs to redundancies in the geometry of the decision boundaries, which was partially supported by their singular value analysis~\cite{moosavi2017universal}. Overall, the authors conclude that there exists a subspace of low dimension in the high-dimensional input space that contains a perturbation vector being somewhat normal to the decision boundary for most images and that the UAP algorithm exploits this to generate the UAP in that subspace. In~\cite{moosavi2017analysis}, UAPs are further interpreted from a geometry perspective in a more fine-grained manner. Specifically, the authors established two models of decision boundaries: (a) flat decision boundaries and (b) curved decision boundaries. The authors showed that \textit{positively curved} decision boundaries provide a better explanation of the existence of UAPs, which is supported by both theoretical analysis and empirical results. Based on this understanding from the geometric perspective, the analysis in~\cite{jetley2018friends} has found that the predictive power and adversarial vulnerability of the studied deep classifier are intertwined, suggesting any gain in robustness must come at the cost of accuracy. 

\begin{figure}[!htbp]
    \centering
    \includegraphics[width=\linewidth]{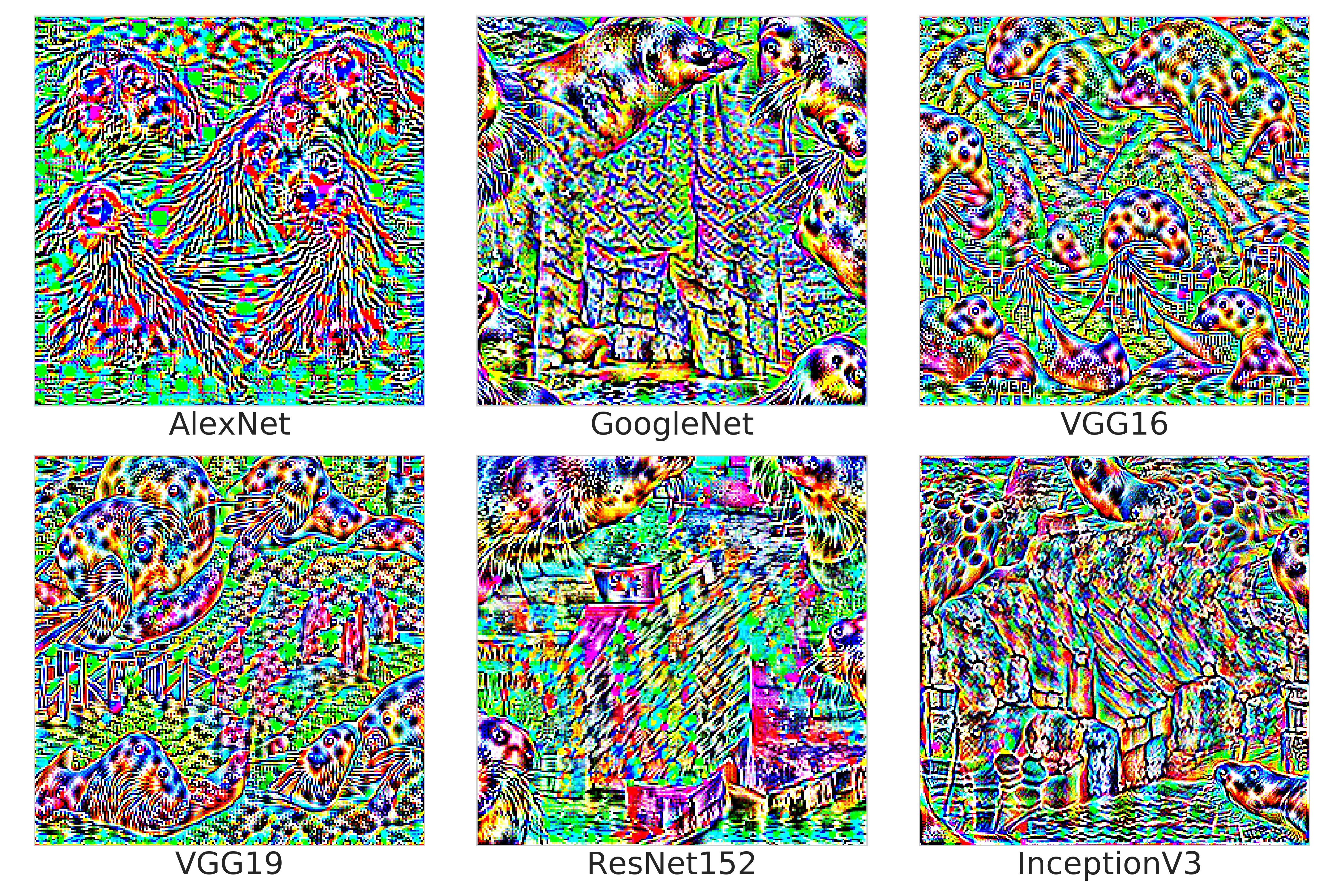}
    \caption{Targeted universal perturbations (target class ``sea lion'') generated with DF-UAP for different network architectures.}
    \label{fig:perturbations_qual}
\end{figure}

\paragraph{Understanding UAPs from a feature perspective.}
Recently, the existence of adversarial examples has been attributed to the non-robust features~\cite{ilyas2019adversarial}. Inspired by this phenomenon, \citeauthor{zhang2020understanding} showed that the UAPs have semantically human-aligned features as shown in Figure~\ref{fig:perturbations_qual}. The chosen target class is ``sea lion", and the human observer can identify the pattern that looks like a sea lion. Specifically, the authors have analyzed the mutual influence of images and perturbations through a Pearson Correlation Coefficient (PCC) analysis and found that UAPs dominate over the images for the model prediction, which is not the case for image-dependent perturbation. As a result, the UAPs can be seen to have independent semantic features and the images behave like noise to them. This is somewhat contrary to the popular belief to only perceive the perturbation as noise to the images. This feature perspective inspires a much more simple yet effective algorithm as discussed in Sec~\ref{sec:attack_methods}. ~\citeauthor{zhang2021universal} have further analyzed the reason why UAP of small magnitude can dominate over images through an investigation of the universal adversarial attack and universal deep hiding~\cite{zhang2020udh}. Frequency is a key factor for the success of both tasks and the reason can be attributed to DNNs being highly sensitive to high-frequency features~\cite{zhang2021universal}. 

\section{Challenges of UAP Attack}
The above discussed UAP algorithms deal with the basic task of universal attacks without taking some underlying challenges into account. Here, we identify three challenges of universal attacks. First, it is not reasonable to get access to the original training dataset that is used for training the target model, thus it is desirable to generate UAPs without the dependence on the original training dataset, \ie data-free UAPs. Second, in practice, access to the target model weights might not be available. Instead, a substitute model that is trained on the same or similar training dataset or the ability to only query the target model might be possible. Thus, black-box universal attacks play also an important role in the area of universal attacks. Lastly, the UAP attacks all images without discrimination, thus causing a serious threat to security-sensitive applications, such as autonomous driving. However, in practice, such an attack also easily catches the attention of the users because samples from all classes are misclassified. Thus, a UAP that can attack the samples in a class-discriminative manner can be more stealthy and might present a more dangerous threat. In the following, we will discuss each of these directions. 

\subsection{Data-Free UAPs}
Despite early works observing that only a subset of the initial training dataset is sufficient to craft UAPs, a data dependence still persists.The issue of crafting data-free UAPs has been addressed by several works. The earliest data-free method was Fast Feature Fool (FFF)~\cite{Mopuri2017datafree}, which generates data-free UAPs by introducing a loss function that maximizes the activations at each layer. This over-firing of the neurons in order to deteriorate the extracted features has been further explored in an extension of their work. The authors demonstrate that the generalizable data-free objective for UAPs (GD-UAP)~\cite{mopuri2018generalizable} can generalize across multiple vision tasks, in particular, image recognition, image segmentation, and depth estimation.
Similar to FFF, the work by~\cite{sam2019crafting} also aims to maximize a certain activation. Specifically, the authors introduce the dilate loss, which maximizes the Euclidean norm of the activation vector before the ReLU layer. We term this approach UAP with dilated loss (UAP-DL).
\citeauthor{mopuri2018ask} craft data-free UAPs by leveraging class-impressions~\cite{mopuri2018ask}. They generate data-free UAPs in a two-stage process. First, class impressions are generated as a substitute for the actual data samples. Class-impressions are generic representations of an object category learned by a model in the input space. Starting from a noisy image, class-impressions are obtained by updating the input noise with the objective to maximize the confidence of the desired class category. In the second stage, a generator is trained to craft UAPs which fool the classifier when added to the class-impression images. During inference, the generated UAP can then be applied to the real images. Instead of generating class impressions, the authors in~\cite{zhang2020understanding} leverage a random proxy dataset, different from the original training dataset, to craft UAPs. 
The authors motivated the use of proxy datasets by their finding that images only behave like noise to the perturbation. To alleviate the need of real images,~\cite{zhang2021jigsaw} further extends by applying jigsaw images to replace proxy dataset.


\subsection{Black-Box UAPs}
\begin{table*}[t]
\centering
    \small
    \scalebox{0.9}{
    \begin{tabular}{ccccccccc}
        \toprule
        Method  & VGG-F & CaffeNet  & GoogleNet  & VGG-16 & VGG-19 &  ResNet152 \\
        \midrule
        UAP~\cite{moosavi2017universal}       & 64.0 & 57.2 & 53.6 & 73.5 & 77.8 & 58.0 \\
        NAG~\cite{mopuri2018nag}              & 67.8 & 67.6 & 74.5 & 80.6 & 83.8 & 65.4 \\
        T-UAP \cite{hashemi2020transferable}  & - & - & - & 84.7 & 94.0 & 36.4 \\
        \midrule
        FFF~\cite{Mopuri2017datafree}         & 39.9 & 38.0 & 30.7 & 38.2 & 43.6 & 26.3 \\
        GD-UAP~\cite{mopuri2018generalizable} & 49.1 & 53.5 & 40.9 & 55.7 & 64.7 & 35.8 \\
        UAP-DL \cite{sam2019crafting}         & -    & -    & 33.7 & 47.5 & 52.0 & 30.4 \\
        AAA~\cite{mopuri2018ask}               & 62.5 & 59.6 & 68.8 & 69.5 & 72.8 & 51.7 \\
        DF-UAP \cite{zhang2020understanding}   & -    & 53.7 & 39.8 & 83.4 & 92.5 & 35.4 \\
        \bottomrule
    \end{tabular}
    }
    \caption{Fooling ratio ($\%$) of various transfer-based attack methods with VGG-19 as the source model. The results are divided into universal attacks with access to the original ImageNet training data (upper) and data-free methods (lower).} 
    \label{tab:black_box_performance_comparison}
\end{table*}

One property of adversarial examples is their transferability, meaning that a perturbation crafted for a source model is also capable of attacking another, unseen model. This is also called a black-box attack since no knowledge about the target model is assumed. The transferability property emphasizes the threat of adversarial examples for the application of Deep Neural Networks in security-critical applications. Transferability is a very active research field for image-dependant attacks. Very few works on UAPs solely focus on the exploration of the transferability properties of UAPs. However, a great portion of the works on UAP report the transferability capabilities of their generated UAPs. 
We summarize the black-box attack capabilities of a few works in Table~\ref{tab:black_box_performance_comparison}. 
Overall Table~\ref{tab:black_box_performance_comparison} shows that in the context of UAPs it is a good rule of thumb that a higher white-box attack rate correlates with a higher black-box capability. Further, we discuss a few works that specifically aim to increase the transferability capabilities of UAPs. The authors of~\cite{li2020regional} investigate the regional homogeneity of UAPs. Their finding suggests that perturbations crafted for models, which are optimized to defend against adversarial examples, show more homogeneous patterns than those crafted for naturally trained models. Therefore, the authors propose regionally homogeneous perturbations (RHP) and showcase their effectiveness against defense models in the transfer setting. 
To achieve more transferable UAPs (T-UAP) \citeauthor{hashemi2020transferable} introduce a new loss that focuses on the adversarial energy in the first layer of source models to work together with the widely used cross-entropy loss to improve its transferability on the target models~\cite{hashemi2020transferable}.
\citeauthor{naseer2019cross} take transferability a step further and shows the existence of domain-invariant adversaries~\cite{naseer2019cross}. The authors show that adversaries learned on Paintings, Cartoons or Medical Images can successfully perturb ImageNet samples to fool the target classifier.
A Decision-based UAP was introduced by~\cite{wu2020decision}. Their decision-based universal adversarial attack (DUAttack) has no access to the internal information of the target models. DUAttack only has access to the hard-label returned by the target models. It utilizes the final inferred label to guide the direction of the perturbation. Specifically, to craft a perturbation with a stripe texture, they apply the orthogonal matrix and iteratively switch the rows of the matrix to determine the location of where the alteration should be applied. Besides, to avoid the altered pixels to offset each other, DUAttack extends its approach with a momentum term, which is commonly used in deep learning, helping to reduce the number of queries. 
The majority of these works discuss the transferability of UAPs in the non-targeted context, meaning that the attack is considered successful if misclassification on the black-box model is achieved. The targeted black-box attack, in which the samples have to be misclassified toward a specific target class is a much more challenging attack scenario and is rarely discussed. Due to the problem setup of UAPs, targeted UAPs can also only be discussed in a more limited scenario, where only one target class can be chosen for all samples since it is unlikely that a single perturbation will be capable to misclassify different samples toward different target classes. The work by~\citeauthor{zhang2020understanding} also considers the data-free targeted UAP case. With relatively low targeted fooling ratios for most networks, but interestingly higher targeted fooling ratios for models from the same architecture families, it emphasizes the difficulty of this attack scenario. Since their attack method can be categorized as a data-free universal attack method, it is considered as the first work to achieve data-free targeted UAPs~\cite{zhang2020understanding}. Improving its performance in the black-box scenario would be an interesting future direction.

\subsection{Class-Discriminative UAPs (CD-UAPs)} All of the previously introduced attack methods attack samples from all classes. The authors of~\cite{zhang2019cd-uap} argue that this obvious misbehavior caused by UAPs might be suspicious to an observer. The authors investigate whether such a UAP exists that only attacks samples from a few classes while limiting the adversarial influence on the remaining classes. Such CD-UAPs would then raise less suspicion since the system under attack would only misbehave when a specific sample from a targeted class would be encountered. By combining separated loss terms for the samples from the non-targeted and targeted samples the authors successfully demonstrate a CD-UAP achieving class discrimination. In an extension to this work, the same group of authors further extends the CD-UAP to a targeted version. The objective of the introduced Double Targeted Attack (DTA)~\cite{benz2020double} is to craft a single perturbation to fool samples of a specific class toward a pre-defined targeted class. Class-wise UAPs have also been explored in~\cite{gupta2019method}. \citeauthor{gupta2019method} propose a data-independent approach to craft CD-UAPs, by exploiting the linearity of the decision boundaries of deep neural networks.




\section{Universal Attack Beyond Classification}
The universal attack against deep classifier has been extended from the image domain to video domain. \citeauthor{li2018adversarial} introduce the first UAP against video recognition~\cite{li2018adversarial}. \citeauthor{chen2019appending} introduce a new variant of a universal attack on videos by appending multiple dummy frames to an arbitrary video clip ~\cite{chen2019appending}. 
We briefly summarize the universal attack in applications beyond image (video) classification.

\subsection{Beyond Classification in the Image Domain} 
\citeauthor{hendrik2017universal} explore how to exploit universal adversarial perturbations against semantic segmentation~\cite{hendrik2017universal}. The authors proposed two successful methods for the attack: the first method is to teach a network to output a desired target segmentation map regardless of the input image; the second method aims to remove target classes from the resulting segmentation map leaving other parts unchanged. \citeauthor{mopuri2018generalizable} show that their proposed data-free GD-UAP also attacks semantic segmentation effectively~\cite{mopuri2018generalizable}. Additionally, the success of GD-UAP for attacking depth estimation is demonstrated. \citeauthor{li2019universal} proposed a universal adversarial perturbation technique against the image-retrieval task~\cite{li2019universal}. The main idea is to attack the point-wise, pair-wise, and list-wise neighborhood relationships. In addition, a coarse-to-fine distillation strategy is also introduced for the black-box attack. Despite good performance on standard benchmarks, the method also extends to real-world systems such as Google Images. 




\subsection{Text Classification}
\citeauthor{wallace2019universal} have introduced universal adversarial triggers for attacking and analyzing natural language processing (NLP)~\cite{wallace2019universal}. Universal adversarial triggers are defined in~\cite{wallace2019universal} as input-agnostic sequences of tokens that can be concatenated to any input from a dataset and consequently result in a specific prediction.
\citeauthor{behjati2019universal} introduce for UAP against text classifier. The UAP for text is defined as a sequence of words that can be added to any input sentence in order and leads to a significant accuracy drop for the text classifier\cite{behjati2019universal}.
The existence of a universal yet small perturbation vector in the embedding space that causes natural text to be misclassified is discovered in~\cite{gao2019universal}. Unlike images for which a UAP of fixed size can be found, the length of the text can change. Thus, the ``universality" has been defined as ``token-agnostic". Specifically, they apply a single perturbation to each token, resulting in different perturbations of flexible sizes at the sequence level. 
The methods introduced in~\cite{wallace2019universal,behjati2019universal,gao2019universal} for attacking text classifier are successful. However, the generated sequence of words do not carry semantic meaning thus can be easily detected by the human. To overcome this drawback, \citeauthor{song2020universal} leverage an adversarially regularized autoencoder (ARAE) for generating natural English phrases that can confuse the text classifier. 

\subsection{Audio Classification}
The existence of UAPs that can fool audio classification architectures for tasks such as speech commands, has been demonstrated in some co-occurring works~\cite{vadillo2019universal,neekhara2019universal}. The algorithms adopted in ~\cite{vadillo2019universal,neekhara2019universal} resemble each other and are inspired by the DeepFool based vanilla UAP algorithm~\cite{moosavi2017universal}. Due to the reasons discussed abovr, such algorithms are often cumbersome and slow. In~\cite{xie2020enabling,li2020universal} UAPs are generated for audio classifier based on generative networks. For example, \citeauthor{xie2020enabling} adopt a Wave-U-Net based fast audio adversarial perturbation generator (FAPG). 
To improve the robustness of the generated UAP against audio, \citeauthor{xie2020real} propose to adopt an acoustic room simulator to estimate the sound distortions~\cite{xie2020real}. Their results show that the proposed acoustic room simulator significantly improves the performance of the UAP. The efficacy of their approach has been demonstrated on a public dataset of 109 speakers. 
Overall, we find that the research in the audio domain is highly influenced by the algorithms developed in the image domain, which is expected because most of the early researches on UAP is exclusively done on the image domain. 

\section{Conclusion}
With a focus on image classification, this survey discusses the recent progress of UAPs for both attack and defense as well as the reason for the existence of UAPs. Additionally, this survey identifies data-dependency, black-box attack, and class-discrimination as three challenges for UAPs and discusses them. This survey also summarizes universal adversarial attacks in a wide range of applications beyond image classification. Overall, the topic of UAP is a fast-evolving field, and our survey can serve as a solid basis for future researches in this field. We believe a joint investigation with data hiding as done in~\cite{zhang2021universal} might be an interesting future direction for providing deeper insight.

\bibliographystyle{ijcai21bst}
\bibliography{bib_mixed1.bib}

\begin{thebibliography}{}

\bibitem[\protect\citeauthoryear{Akhtar and Mian}{2018}]{akhtar2018threat}
Naveed Akhtar and Ajmal Mian.
\newblock Threat of adversarial attacks on deep learning in computer vision: A
  survey.
\newblock {\em IEEE Access}, 2018.

\bibitem[\protect\citeauthoryear{Akhtar \bgroup \em et al.\egroup
  }{2018}]{akhtar2018defense}
Naveed Akhtar, Jian Liu, and Ajmal Mian.
\newblock Defense against universal adversarial perturbations.
\newblock In {\em CVPR}, 2018.

\bibitem[\protect\citeauthoryear{Behjati \bgroup \em et al.\egroup
  }{2019}]{behjati2019universal}
Melika Behjati, Seyed-Mohsen Moosavi-Dezfooli, Mahdieh~Soleymani Baghshah, and
  Pascal Frossard.
\newblock Universal adversarial attacks on text classifiers.
\newblock In {\em ICASSP}. IEEE, 2019.

\bibitem[\protect\citeauthoryear{Benz \bgroup \em et al.\egroup
  }{2020}]{benz2020double}
Philipp Benz, Chaoning Zhang, Tooba Imtiaz, and In~So Kweon.
\newblock Double targeted universal adversarial perturbations.
\newblock In {\em ACCV}, 2020.

\bibitem[\protect\citeauthoryear{Benz \bgroup \em et al.\egroup
  }{2021}]{benz2021universal}
Philipp Benz, Chaoning Zhang, Adil Karjauv, and In~So Kweon.
\newblock Universal adversarial training with class-wise perturbations.
\newblock {\em ICME}, 2021.

\bibitem[\protect\citeauthoryear{Borkar \bgroup \em et al.\egroup
  }{2020}]{borkar2020defending}
Tejas Borkar, Felix Heide, and Lina Karam.
\newblock Defending against universal attacks through selective feature
  regeneration.
\newblock In {\em CVPR}, 2020.

\bibitem[\protect\citeauthoryear{Carlini and Wagner}{2017}]{carlini2017towards}
Nicholas Carlini and David Wagner.
\newblock Towards evaluating the robustness of neural networks.
\newblock In {\em Symposium on Security and Privacy (SP)}, 2017.

\bibitem[\protect\citeauthoryear{Chen \bgroup \em et al.\egroup
  }{2019}]{chen2019appending}
Zhikai Chen, Lingxi Xie, Shanmin Pang, Yong He, and Qi~Tian.
\newblock Appending adversarial frames for universal video attack.
\newblock {\em arXiv preprint arXiv:1912.04538}, 2019.

\bibitem[\protect\citeauthoryear{Dai and Shu}{2019}]{dai2019fast}
Jiazhu Dai and Le~Shu.
\newblock Fast-uap: Algorithm for speeding up universal adversarial
  perturbation generation with orientation of perturbation vectors.
\newblock {\em arXiv preprint arXiv:1911.01172}, 2019.

\bibitem[\protect\citeauthoryear{Fawzi \bgroup \em et al.\egroup
  }{2016}]{fawzi2016robustness}
Alhussein Fawzi, Seyed-Mohsen Moosavi-Dezfooli, and Pascal Frossard.
\newblock Robustness of classifiers: from adversarial to random noise.
\newblock In {\em NeurIPS}, 2016.

\bibitem[\protect\citeauthoryear{Gao and Oates}{2019}]{gao2019universal}
Hang Gao and Tim Oates.
\newblock Universal adversarial perturbation for text classification.
\newblock {\em arXiv preprint arXiv:1910.04618}, 2019.

\bibitem[\protect\citeauthoryear{Goodfellow \bgroup \em et al.\egroup
  }{2014}]{goodfellow2014generative}
Ian Goodfellow, Jean Pouget-Abadie, Mehdi Mirza, Bing Xu, David Warde-Farley,
  Sherjil Ozair, Aaron Courville, and Yoshua Bengio.
\newblock Generative adversarial nets.
\newblock In {\em NeurIPS}, 2014.

\bibitem[\protect\citeauthoryear{Goodfellow \bgroup \em et al.\egroup
  }{2015}]{goodfellow2014explaining}
Ian~J Goodfellow, Jonathon Shlens, and Christian Szegedy.
\newblock Explaining and harnessing adversarial examples.
\newblock In {\em ICLR}, 2015.

\bibitem[\protect\citeauthoryear{Gupta \bgroup \em et al.\egroup
  }{2019}]{gupta2019method}
Tejus Gupta, Abhishek Sinha, Nupur Kumari, Mayank Singh, and Balaji
  Krishnamurthy.
\newblock A method for computing class-wise universal adversarial
  perturbations.
\newblock {\em arXiv preprint arXiv:1912.00466}, 2019.

\bibitem[\protect\citeauthoryear{Hashemi \bgroup \em et al.\egroup
  }{2020}]{hashemi2020transferable}
Atiye~Sadat Hashemi, Andreas B{\"a}r, Saeed Mozaffari, and Tim Fingscheidt.
\newblock Transferable universal adversarial perturbations using generative
  models.
\newblock {\em arXiv preprint arXiv:2010.14919}, 2020.

\bibitem[\protect\citeauthoryear{Hayes and Danezis}{2018}]{hayes2018learning}
Jamie Hayes and George Danezis.
\newblock Learning universal adversarial perturbations with generative models.
\newblock In {\em IEEE Security and Privacy Workshops (SPW)}, 2018.

\bibitem[\protect\citeauthoryear{Hendrik~Metzen \bgroup \em et al.\egroup
  }{2017}]{hendrik2017universal}
Jan Hendrik~Metzen, Mummadi Chaithanya~Kumar, Thomas Brox, and Volker Fischer.
\newblock Universal adversarial perturbations against semantic image
  segmentation.
\newblock In {\em ICCV}, 2017.

\bibitem[\protect\citeauthoryear{Ilyas \bgroup \em et al.\egroup
  }{2019}]{ilyas2019adversarial}
Andrew Ilyas, Shibani Santurkar, Dimitris Tsipras, Logan Engstrom, Brandon
  Tran, and Aleksander Madry.
\newblock Adversarial examples are not bugs, they are features.
\newblock In {\em NeurIPS}, 2019.

\bibitem[\protect\citeauthoryear{Jetley \bgroup \em et al.\egroup
  }{2018}]{jetley2018friends}
Saumya Jetley, Nicholas Lord, and Philip Torr.
\newblock With friends like these, who needs adversaries?
\newblock In {\em NeurIPS}, 2018.

\bibitem[\protect\citeauthoryear{Khrulkov and
  Oseledets}{2018}]{khrulkov2018art}
Valentin Khrulkov and Ivan Oseledets.
\newblock Art of singular vectors and universal adversarial perturbations.
\newblock In {\em CVPR}, 2018.

\bibitem[\protect\citeauthoryear{Kurakin \bgroup \em et al.\egroup
  }{2017}]{kurakin2016adversarial}
Alexey Kurakin, Ian Goodfellow, and Samy Bengio.
\newblock Adversarial machine learning at scale.
\newblock In {\em ICLR}, 2017.

\bibitem[\protect\citeauthoryear{Li \bgroup \em et al.\egroup
  }{2018}]{li2018adversarial}
Shasha Li, Ajaya Neupane, Sujoy Paul, Chengyu Song, Srikanth~V Krishnamurthy,
  Amit K~Roy Chowdhury, and Ananthram Swami.
\newblock Adversarial perturbations against real-time video classification
  systems.
\newblock {\em arXiv preprint arXiv:1807.00458}, 2018.

\bibitem[\protect\citeauthoryear{Li \bgroup \em et al.\egroup
  }{2019}]{li2019universal}
Jie Li, Rongrong Ji, Hong Liu, Xiaopeng Hong, Yue Gao, and Qi~Tian.
\newblock Universal perturbation attack against image retrieval.
\newblock In {\em ICCV}, 2019.

\bibitem[\protect\citeauthoryear{Li \bgroup \em et al.\egroup
  }{2020a}]{li2020universal}
Jiguo Li, Xinfeng Zhang, Chuanmin Jia, Jizheng Xu, Li~Zhang, Yue Wang, Siwei
  Ma, and Wen Gao.
\newblock Universal adversarial perturbations generative network for speaker
  recognition.
\newblock In {\em ICME}, 2020.

\bibitem[\protect\citeauthoryear{Li \bgroup \em et al.\egroup
  }{2020b}]{li2020regional}
Yingwei Li, Song Bai, Cihang Xie, Zhenyu Liao, Xiaohui Shen, and Alan~L Yuille.
\newblock Regional homogeneity: Towards learning transferable universal
  adversarial perturbations against defenses.
\newblock In {\em ECCV}, 2020.

\bibitem[\protect\citeauthoryear{Liu \bgroup \em et al.\egroup
  }{2019}]{liu2019universal}
Hong Liu, Rongrong Ji, Jie Li, Baochang Zhang, Yue Gao, Yongjian Wu, and Feiyue
  Huang.
\newblock Universal adversarial perturbation via prior driven uncertainty
  approximation.
\newblock In {\em ICCV}, 2019.

\bibitem[\protect\citeauthoryear{Madry \bgroup \em et al.\egroup
  }{2018}]{madry2017towards}
Aleksander Madry, Aleksandar Makelov, Ludwig Schmidt, Dimitris Tsipras, and
  Adrian Vladu.
\newblock Towards deep learning models resistant to adversarial attacks.
\newblock In {\em ICLR}, 2018.

\bibitem[\protect\citeauthoryear{Moosavi-Dezfooli \bgroup \em et al.\egroup
  }{2016}]{moosavi2016deepfool}
Seyed-Mohsen Moosavi-Dezfooli, Alhussein Fawzi, and Pascal Frossard.
\newblock Deepfool: a simple and accurate method to fool deep neural networks.
\newblock In {\em CVPR}, 2016.

\bibitem[\protect\citeauthoryear{Moosavi-Dezfooli \bgroup \em et al.\egroup
  }{2017a}]{moosavi2017universal}
Seyed-Mohsen Moosavi-Dezfooli, Alhussein Fawzi, Omar Fawzi, and Pascal
  Frossard.
\newblock Universal adversarial perturbations.
\newblock In {\em CVPR}, 2017.

\bibitem[\protect\citeauthoryear{Moosavi-Dezfooli \bgroup \em et al.\egroup
  }{2017b}]{moosavi2017analysis}
Seyed-Mohsen Moosavi-Dezfooli, Alhussein Fawzi, Omar Fawzi, Pascal Frossard,
  and Stefano Soatto.
\newblock Analysis of universal adversarial perturbations.
\newblock {\em arXiv preprint arXiv:1705.09554}, 2017.

\bibitem[\protect\citeauthoryear{Mopuri \bgroup \em et al.\egroup
  }{2017}]{Mopuri2017datafree}
Konda~Reddy Mopuri, Utsav Garg, and R.~Venkatesh Babu.
\newblock Fast feature fool: A data independent approach to universal
  adversarial perturbations.
\newblock In {\em BMVC}, 2017.

\bibitem[\protect\citeauthoryear{Mopuri \bgroup \em et al.\egroup
  }{2018a}]{mopuri2018generalizable}
Konda~Reddy Mopuri, Aditya Ganeshan, and Venkatesh~Babu Radhakrishnan.
\newblock Generalizable data-free objective for crafting universal adversarial
  perturbations.
\newblock {\em TPAMI}, 2018.

\bibitem[\protect\citeauthoryear{Mopuri \bgroup \em et al.\egroup
  }{2018b}]{mopuri2018nag}
Konda~Reddy Mopuri, Utkarsh Ojha, Utsav Garg, and R.~Venkatesh Babu.
\newblock Nag: Network for adversary generation.
\newblock In {\em CVPR}, 2018.

\bibitem[\protect\citeauthoryear{Mopuri \bgroup \em et al.\egroup
  }{2018c}]{mopuri2018ask}
Konda~Reddy Mopuri, Phani~Krishna Uppala, and R.~Venkatesh Babu.
\newblock Ask, acquire, and attack: Data-free uap generation using class
  impressions.
\newblock In {\em ECCV}, 2018.

\bibitem[\protect\citeauthoryear{Mummadi \bgroup \em et al.\egroup
  }{2019}]{mummadi2019defending}
Chaithanya~Kumar Mummadi, Thomas Brox, and Jan~Hendrik Metzen.
\newblock Defending against universal perturbations with shared adversarial
  training.
\newblock In {\em ICCV}, 2019.

\bibitem[\protect\citeauthoryear{Naseer \bgroup \em et al.\egroup
  }{2019}]{naseer2019cross}
Muhammad~Muzammal Naseer, Salman~H Khan, Muhammad~Haris Khan, Fahad
  Shahbaz~Khan, and Fatih Porikli.
\newblock Cross-domain transferability of adversarial perturbations.
\newblock In {\em NeurIPS}, 2019.

\bibitem[\protect\citeauthoryear{Neekhara \bgroup \em et al.\egroup
  }{2019}]{neekhara2019universal}
Paarth Neekhara, Shehzeen Hussain, Prakhar Pandey, Shlomo Dubnov, Julian
  McAuley, and Farinaz Koushanfar.
\newblock Universal adversarial perturbations for speech recognition systems.
\newblock {\em arXiv preprint arXiv:1905.03828}, 2019.

\bibitem[\protect\citeauthoryear{Perolat \bgroup \em et al.\egroup
  }{2018}]{perolat2018playing}
Julien Perolat, Mateusz Malinowski, Bilal Piot, and Olivier Pietquin.
\newblock Playing the game of universal adversarial perturbations.
\newblock {\em arXiv preprint arXiv:1809.07802}, 2018.

\bibitem[\protect\citeauthoryear{Poursaeed \bgroup \em et al.\egroup
  }{2018}]{poursaeed2018generative}
Omid Poursaeed, Isay Katsman, Bicheng Gao, and Serge Belongie.
\newblock Generative adversarial perturbations.
\newblock In {\em CVPR}, 2018.

\bibitem[\protect\citeauthoryear{Sam \bgroup \em et al.\egroup
  }{2019}]{sam2019crafting}
Deepak~Babu Sam, KA~Sudharsan, Venkatesh~Babu Radhakrishnan, et~al.
\newblock Crafting data-free universal adversaries with dilate loss.
\newblock 2019.

\bibitem[\protect\citeauthoryear{Schmidt \bgroup \em et al.\egroup
  }{2018}]{schmidt2018adversarially}
Ludwig Schmidt, Shibani Santurkar, Dimitris Tsipras, Kunal Talwar, and
  Aleksander Madry.
\newblock Adversarially robust generalization requires more data.
\newblock In {\em NeurIPS}, 2018.

\bibitem[\protect\citeauthoryear{Shafahi \bgroup \em et al.\egroup
  }{2019a}]{shafahi2018adversarial}
Ali Shafahi, W.~Ronny Huang, Christoph Studer, Soheil Feizi, and Tom Goldstein.
\newblock Are adversarial examples inevitable?
\newblock In {\em ICLR}, 2019.

\bibitem[\protect\citeauthoryear{Shafahi \bgroup \em et al.\egroup
  }{2019b}]{shafahi2019adversarial}
Ali Shafahi, Mahyar Najibi, Mohammad~Amin Ghiasi, Zheng Xu, John Dickerson,
  Christoph Studer, Larry~S Davis, Gavin Taylor, and Tom Goldstein.
\newblock Adversarial training for free!
\newblock In {\em NeurIPS}, 2019.

\bibitem[\protect\citeauthoryear{Shafahi \bgroup \em et al.\egroup
  }{2020}]{shafahi2020universal}
Ali Shafahi, Mahyar Najibi, Zheng Xu, John~P Dickerson, Larry~S Davis, and Tom
  Goldstein.
\newblock Universal adversarial training.
\newblock In {\em AAAI}, 2020.

\bibitem[\protect\citeauthoryear{Song \bgroup \em et al.\egroup
  }{2020}]{song2020universal}
Liwei Song, Xinwei Yu, Hsuan-Tung Peng, and Karthik Narasimhan.
\newblock Universal adversarial attacks with natural triggers for text
  classification.
\newblock {\em arXiv preprint arXiv:2005.00174}, 2020.

\bibitem[\protect\citeauthoryear{Szegedy \bgroup \em et al.\egroup
  }{2013}]{szegedy2013intriguing}
Christian Szegedy, Wojciech Zaremba, Ilya Sutskever, Joan Bruna, Dumitru Erhan,
  Ian Goodfellow, and Rob Fergus.
\newblock Intriguing properties of neural networks.
\newblock {\em arXiv preprint arXiv:1312.6199}, 2013.

\bibitem[\protect\citeauthoryear{Vadillo and
  Santana}{2019}]{vadillo2019universal}
Jon Vadillo and Roberto Santana.
\newblock Universal adversarial examples in speech command classification.
\newblock {\em arXiv preprint arXiv:1911.10182}, 2019.

\bibitem[\protect\citeauthoryear{Wallace \bgroup \em et al.\egroup
  }{2019}]{wallace2019universal}
Eric Wallace, Shi Feng, Nikhil Kandpal, Matt Gardner, and Sameer Singh.
\newblock Universal adversarial triggers for attacking and analyzing nlp.
\newblock {\em arXiv preprint arXiv:1908.07125}, 2019.

\bibitem[\protect\citeauthoryear{Wong \bgroup \em et al.\egroup
  }{2020}]{wong2020fast}
Eric Wong, Leslie Rice, and J~Zico Kolter.
\newblock Fast is better than free: Revisiting adversarial training.
\newblock {\em ICLR}, 2020.

\bibitem[\protect\citeauthoryear{Wu \bgroup \em et al.\egroup
  }{2020}]{wu2020decision}
Jing Wu, Mingyi Zhou, Shuaicheng Liu, Yipeng Liu, and Ce~Zhu.
\newblock Decision-based universal adversarial attack.
\newblock {\em arXiv preprint arXiv:2009.07024}, 2020.

\bibitem[\protect\citeauthoryear{Xie \bgroup \em et al.\egroup
  }{2020a}]{xie2020enabling}
Yi~Xie, Zhuohang Li, Cong Shi, Jian Liu, Yingying Chen, and Bo~Yuan.
\newblock Enabling fast and universal audio adversarial attack using generative
  model.
\newblock {\em arXiv preprint arXiv:2004.12261}, 2020.

\bibitem[\protect\citeauthoryear{Xie \bgroup \em et al.\egroup
  }{2020b}]{xie2020real}
Yi~Xie, Cong Shi, Zhuohang Li, Jian Liu, Yingying Chen, and Bo~Yuan.
\newblock Real-time, universal, and robust adversarial attacks against speaker
  recognition systems.
\newblock In {\em ICASSP}. IEEE, 2020.

\bibitem[\protect\citeauthoryear{Zhang \bgroup \em et al.\egroup
  }{2019}]{zhang2019you}
Dinghuai Zhang, Tianyuan Zhang, Yiping Lu, Zhanxing Zhu, and Bin Dong.
\newblock You only propagate once: Accelerating adversarial training via
  maximal principle.
\newblock In {\em NeurIPS}, 2019.

\bibitem[\protect\citeauthoryear{Zhang \bgroup \em et al.\egroup
  }{2020a}]{zhang2019cd-uap}
Chaoning Zhang, Philipp Benz, Tooba Imtiaz, and In-So Kweon.
\newblock Cd-uap: Class discriminative universal adversarial perturbation.
\newblock In {\em AAAI}, 2020.

\bibitem[\protect\citeauthoryear{Zhang \bgroup \em et al.\egroup
  }{2020b}]{zhang2020understanding}
Chaoning Zhang, Philipp Benz, Tooba Imtiaz, and In-So Kweon.
\newblock Understanding adversarial examples from the mutual influence of
  images and perturbations.
\newblock In {\em CVPR}, 2020.

\bibitem[\protect\citeauthoryear{Zhang \bgroup \em et al.\egroup
  }{2020c}]{zhang2020udh}
Chaoning Zhang, Philipp Benz, Adil Karjauv, Geng Sun, and In~Kweon.
\newblock Udh: Universal deep hiding for steganography, watermarking, and light
  field messaging.
\newblock {\em NeurIPS}, 2020.

\bibitem[\protect\citeauthoryear{Zhang \bgroup \em et al.\egroup
  }{2021a}]{zhang2021jigsaw}
Chaoning Zhang, Philipp Benz, Adil Karjauv, Jae~Won Cho, and In~So Kweon.
\newblock Towards data-free universal adversarial perturbations with artificial
  images.
\newblock {\em RobustML workshop at ICLR2021}, 2021.

\bibitem[\protect\citeauthoryear{Zhang \bgroup \em et al.\egroup
  }{2021b}]{zhang2021universal}
Chaoning Zhang, Philipp Benz, Adil Karjauv, and In~So Kweon.
\newblock Universal adversarial perturbations through the lens of deep
  steganography: Towards a fourier perspective.
\newblock {\em AAAI}, 2021.

\end{thebibliography}

\end{document}